\documentclass[runningheads]{llncs}
\usepackage{hyperref}
\usepackage{booktabs}
\usepackage{graphicx}
\usepackage{longtable}
\usepackage{amsmath}
\usepackage{placeins}
\usepackage{amssymb}
\begin{document}
\title{Cycle Diffusion Model for Counterfactual Image Generation}
% \author{Fangrui Huang\inst{1} \and
% Alan Wang\inst{1} \and
% Binxu Li\inst{1} \and
% Bailey Trang \and
% Ridvan Yesiloglu\inst{1} \and
% Tianyu Hua\inst{1} \and
% Wei Peng\inst{1} \and
% Ehsan Adeli\inst{1}
% }
\author{Fangrui Huang \and
Alan Wang \and
Binxu Li \and
Bailey Trang \and
Ridvan Yesiloglu \and 
\\Tianyu Hua \and
Wei Peng \and
Ehsan Adeli
}
\authorrunning{F. Huang et al.}
\institute{Stanford University, Stanford CA 94305, USA \\
 \email{fangruih@stanford.edu}\\}
%  \and
\maketitle              % typeset the header of the contribution
\begin{abstract}
Deep generative models have demonstrated remarkable success in medical image synthesis.
However, ensuring conditioning faithfulness and high-quality synthetic images for direct or
counterfactual generation remains a challenge. In this work, we introduce a cycle training
framework to fine-tune diffusion models for improved conditioning adherence and enhanced
synthetic image realism. Our approach, Cycle Diffusion Model (CDM), enforces consistency between generated and original images by incorporating cycle constraints, enabling
more reliable direct and counterfactual generation. Experiments on a combined 3D brain
MRI dataset (from ABCD, HCP aging \& young adults, ADNI, and PPMI) show that our
method improves conditioning accuracy and enhances image quality as measured by FID
and SSIM. The results suggest that the cycle strategy used in CDM can be an effective
method for refining diffusion-based medical image generation, with applications
in data augmentation, counterfactual, and disease progression modeling.

\end{abstract}
\section{Introduction}
Medical image synthesis plays a crucial role in advancing the understanding of human anatomy and improving diagnostic tools in healthcare~\cite{KHALIFA2024100146,yi2019ganinmedicalimaging}. 
In particular, the generation of 3D brain MRIs with precise control over demographic factors such as age and sex has significant potential for various applications, including personalized medicine \cite{berko2023mri,dalessandris2024personalized}, disease progression modeling \cite{puglisi2024enhancing}, and augmenting training datasets for AI models~\cite{pombo2023equitablemodelling}. 
Popular generative models, such as Generative Adversarial Networks (GANs)~\cite{goodfellow2014generativeadversarialnetworks} and Latent Diffusion Models (LDMs)~\cite{rombach2022highresolution}, have shown success in generating realistic medical images, but often struggle with adhering to conditioning variables or producing diverse yet anatomically accurate samples~\cite{dhinagar2024counterfactual}.

In this paper, we propose a novel cycle diffusion model (CDM) for 3D brain MRI generation, which improves upon existing diffusion models by leveraging a cycle-consistent training procedure that enforces accurate conditioning adherence while enhancing the realism of generated images. 
Specifically, our framework focuses on generating 3D brain MRI volumes, with predefined conditions (e.g., both age and sex), and it is designed to support both direct generation (where a specific age and sex are input) and counterfactual generation (where a brain MRI is transformed to reflect a different condition).

To evaluate the success of these generative tasks, we perform a comprehensive analysis of age and sex prediction accuracy on synthetic samples. 
Additionally, we compare our approach to existing generative models using quantitative metrics such as Frechet Inception Distance (FID)~\cite{heusel2018ganstrainedtimescaleupdate} and Multi-Scale Structural Similarity (MS-SSIM)~\cite{wang2003msssim}, along with qualitative assessments of the morphological realism of the generated images. 
Our results demonstrate that CDM outperforms baseline approaches in terms of image quality, diversity, and conditioning adherence while maintaining anatomical accuracy and structural integrity in both direct and counterfactual generation tasks.

\begin{figure}[t]
\includegraphics[width=\textwidth]{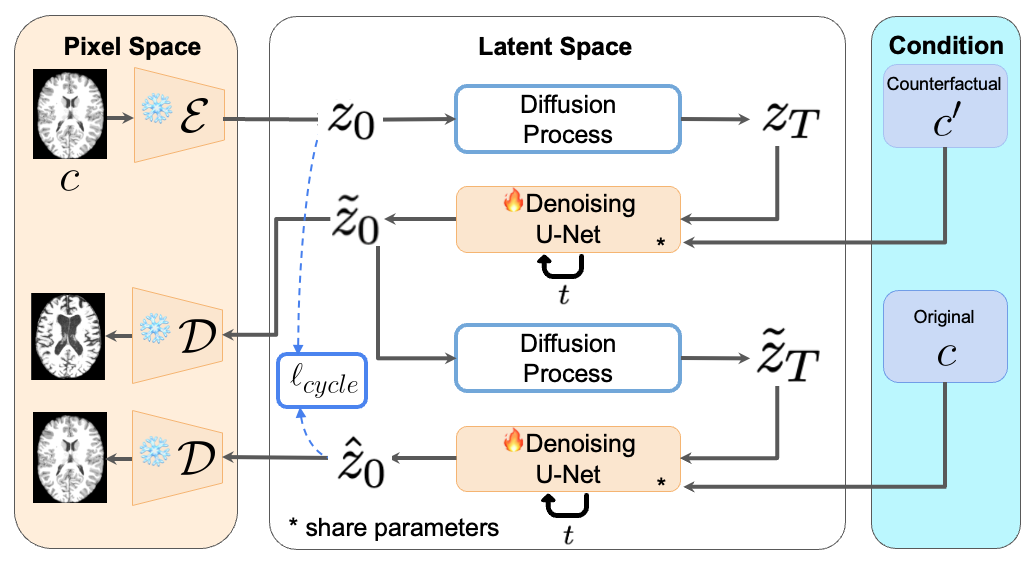}
\caption{
Graphical depiction of the proposed method.
Our cycle diffusion model (CDM) performs a generative denoising process in the latent space in two directions.
The counterfactual direction produces a counterfactual latent $\tilde{z}_0$ that is conditioned on the counterfactual condition $c'$.  
The factual direction produces a factual latent $\hat{z}_0$ that is conditioned on the original condition $c$. 
In addition to the denoising loss, a cycle-consistency loss is optimized which minimizes the distance between $z_0$ and $\hat{z_0}$. 
% By comparing the predicted reconstruction of $z_0$ to the original, the model can learn to preserve global structure after condition changes. 
% $\tilde{z_0}$ and $\hat{z_0}$ are obtained by inverting Equation \ref{eq:zt}. 
} \label{model}
\end{figure}

\section{Related Work}
\noindent\textbf{Medical Image Generation using Generative Models:}
Generative models play a vital role in medical imaging by addressing data scarcity, high annotation costs, and model robustness. GANs~\cite{goodfellow2014generativeadversarialnetworks} and VAEs~\cite{kingma2013auto} are widely used to synthesize realistic medical images, aiding in data augmentation for machine learning tasks~\cite{wang2020spatial,pombo2023equitablemodelling,peng2024latent}. A key challenge is conditioning adherence—ensuring generated images align with attributes like age, sex, or disease state—without compromising anatomical realism, where GAN approaches struggle with \cite{peng2024generative}.

\noindent\textbf{Diffusion Models in Medical Imaging:}
Diffusion models (DMs) have recently gained attention in medical imaging for their ability to generate high-quality, stable synthetic images. Latent Diffusion Models (LDMs)~\cite{rombach2022highresolution}, in particular, have shown promise in 3D imaging tasks like MRI synthesis. These models generate images by learning to reverse a noise-adding process, and are able to generate high-quality synthetic images. However, DMs still face challenges with precise conditioning on demographic and disease-related factors~\cite{peng2024generative}. Recent advances, such as metadata-conditioned models like BrainSynth~\cite{peng2024metadata}, improve conditioning adherence by leveraging subject-level metadata during synthesis, enhancing visual quality and clinical relevance.

DMs are especially valuable for data augmentation, enabling the generation of diverse, clinically relevant images to improve model generalization—crucial for rare diseases and underrepresented populations. Approaches like MedSyn~\cite{10566053} support text-guided, anatomy-aware 3D CT synthesis, while counterfactual generation techniques aid in interpretable disease effect detection~\cite{dhinagar2024counterfactual}. Further developments include shape-conditioned MRI generation~\cite{bongratz20253d}, highlighting the growing role of DMs in enhancing clinical datasets.

\noindent\textbf{Cycle Consistency in Medical Image Generation:}
% Ensuring accurate conditioning is essential in medical imaging, where the reliability of synthetic data is critical for downstream clinical use.
Cycle-consistency, popularized by models like CycleGAN~\cite{zhu2017unpaired}, has proven effective in preserving structural integrity during image-to-image translation by enforcing a reversible transformation.
Recent work has extended this concept to diffusion models~\cite{wu2022unifying,xu2024cyclenetrethinkingcycleconsistency} in natural images, enabling better control of conditioning in natural image editing. Cycle consistency shows potential in improving conditioning adherence in medical imaging generation tasks, but needs to be experimented with. 

% By incorporating cycle consistency, diffusion models have the potential to generate more condition-specific medical imaging data or improve identity preservation for medical imagining translation tasks. This integration improves the clinical relevance of synthetic datasets.
% \subsection{Counterfactual Generation and Medical Image Editing}
% Counterfactual generation is a critical aspect of medical image generation, especially in the context of disease progression modeling and understanding the effect of different clinical interventions. DDIM-based sampling~\cite{Song2020DenoisingDI} has been used to generate counterfactuals, where medical images are edited to reflect different conditions. This is useful for generating synthetic images that can simulate various scenarios, such as how the brain might look in the future with neurodegenerative changes, or how a treatment might reverse or mitigate those changes.

% By utilizing counterfactual generation in medical imaging, we can simulate disease progression, treatment response, and even explore the potential effects of different interventions. These capabilities open up new possibilities for both research and clinical practice, allowing for a deeper understanding of disease mechanisms and offering new tools for personalized treatment strategies.

\noindent\textbf{Summary of Our Contributions:}
Our work introduces CDM, which combines cycle consistency with diffusion-based generative modeling for medical image synthesis. This model improves conditioning adherence through the training strategy that incorporates cycle-consistent regularization as introduced in~\ref{sec:methods}. Our method shows significant promise for applications in medical image augmentation, disease progression modeling, and counterfactual generation, providing a valuable tool for researchers and clinicians alike.

\begin{figure}[t]
\includegraphics[width=\textwidth]{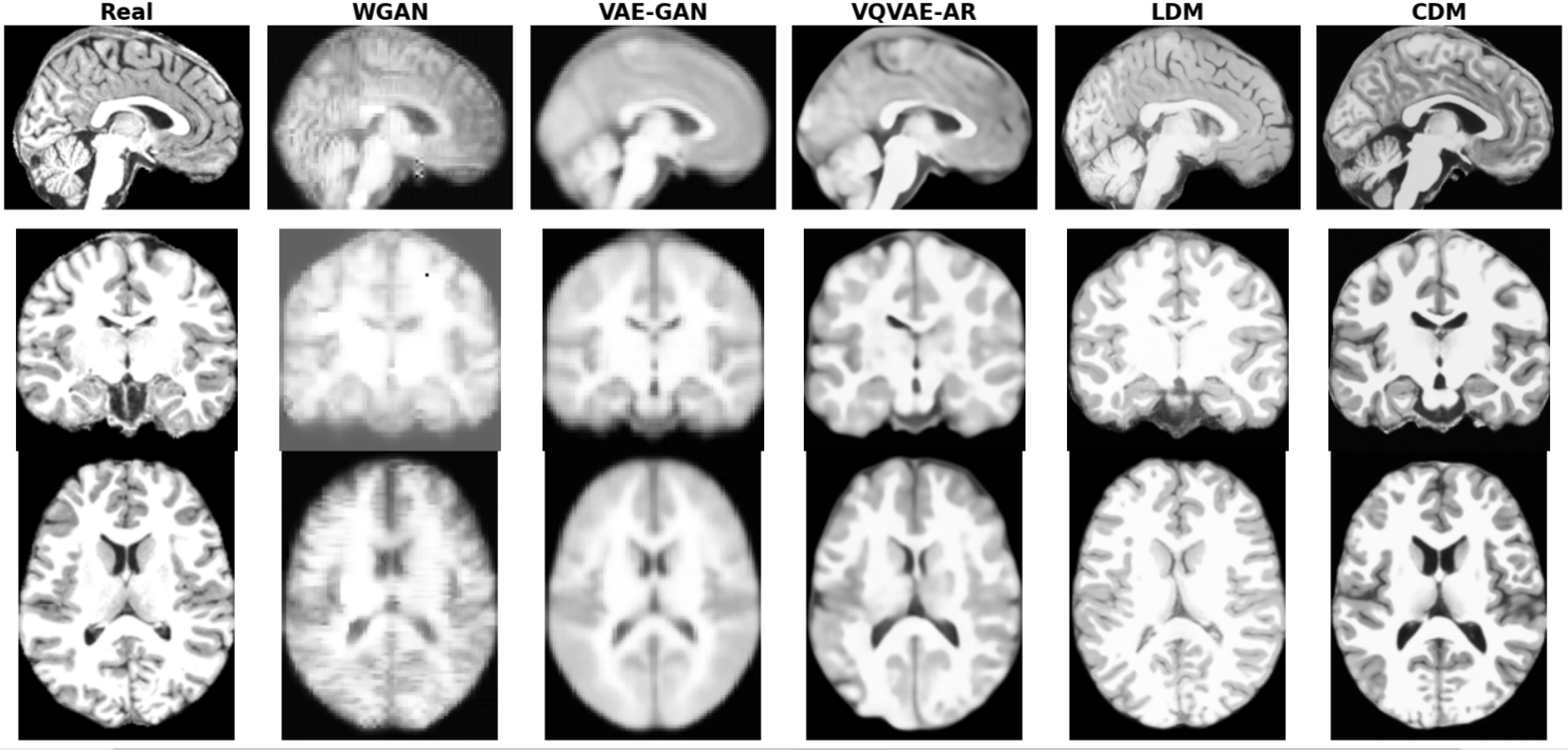}
\caption{Representative real and synthetic samples for all models across the three planes are shown. The comparison includes the following methods, trained with the same experimental conditions as our method: VAE-GAN\cite{larsen2016autoencoding}, $\alpha$-GAN \cite{kwon2019braingan}, VQVAE-AR \cite{van2017neural,tudosiu2022autoregressive}, and LDM \cite{rombach2022highresolution,pinaya2022brainimaginggenerationlatent}. The selected subject is a 13-year-old male. We observe clearer ventricles and more detailed folding patterns on the cortex in CDM compared to baselines.} \label{direct visualization}
\end{figure}
\section{Methods}
\label{sec:methods}
%\subsection{Motivation}
Generating counterfactual brain images conditioned on demographic attributes such as age and sex is a compelling but challenging task. Counterfactuals can support the analysis of structural variability across populations, facilitate longitudinal predictions, or simulate hypothetical interventions. However, in the absence of paired brain scans under different demographic conditions for the same individual, this becomes a fundamentally ill-posed problem.

To address this, we draw inspiration from cycle-consistency frameworks that have shown remarkable success in similar unpaired translation tasks. In particular, CycleGAN \cite{zhu2017unpaired} and CycleNet \cite{zhang2023cyclenet} demonstrate that enforcing consistency between forward and backward mappings enables models to learn structure-preserving transformations, even in the absence of paired data.

In our setting, this idea is especially well-suited: brain anatomy is highly individualized, and small structural differences (e.g., between sexes or across age ranges) must be modeled without distorting the underlying subject-specific features.  By incorporating cycle consistency, we explicitly constrain the model to make only minimal, semantically valid changes that can be reversed—helping preserve anatomical fidelity.

%\subsection{Overview}
Our proposed method (Figure \ref{model}) leverages this principle via a Cycle Diffusion Model (CDM) that performs bidirectional training using counterfactual and factual generation steps.
% Similar to LDM training, CDM is trained using unpaired brain images, conditioned on demographic attributes such as age and sex. 
Instead of a single generation direction, during training, we perform a ``counterfactual'' generation process followed by a ``factual'' generation process.
In addition to minimizing the same LDM denoising loss as in typical LDM training, we enforce an additional cycle consistency loss ($\ell_\text{cycle}$) that encourages the output of the factual generation to be as close as possible to the original image.
Intuitively, the additional cycle-consistency regularization ensures that the generated images adhere to the input conditions while biasing the model to make the minimal changes necessary with respect to the original image to satisfy the cycle-consistency constraint.

% \subsection{Loss functions}
% Our loss functions are as follows. Let $z=z_0$ denote a latent embedding and $c$ denote its corresponding condition.
% Let $c' \neq c$ define a counterfactual condition. As a result, the cycle-consistency loss is defined as:

% \begin{equation}
%    \ell_{\text{cycle}}(z_0, t, c, c') = ||z_0 - \hat{z}_0(\hat{z}_0(z_t, t, c'), t, c)||.
% \label{eq:cycle_loss}
% \end{equation}
% Note that the input condition for the "factual" process is the same as the original condition of $z$.
% The overall loss is the LDM loss in both the counterfactual and factual directions, plus the cycle-consistency loss:
% \begin{equation}
% \mathcal{L}_\text{CDM} = \mathbb{E}_{z, \epsilon, t} \left[\ell_{\text{LDM}}(\epsilon, z_t, t, c') + \ell_{\text{LDM}}(\epsilon, \tilde{z}_t, t, c) + \lambda \ell_\text{cycle}(z, t, c, c') \right].
% \label{eq:composite_loss}
% \end{equation}
% where $\tilde{z} = \hat{z}_0(z_t, t, c')$ and $\lambda$ is a hyperparameter.
\subsection{Loss Functions}

Our model builds upon the standard Latent Diffusion Model (LDM) training objective by introducing a cycle-consistency loss that enables structure-preserving counterfactual generation. 

Let $z=z_0$ denote a latent embedding of the input brain image under condition $c$, and let $c' \neq c$ represent a counterfactual condition.
We define a forward process that generates a counterfactual latent sample $\tilde{z}_0 = \hat{z}_0(z_t, t, c')$, followed by a backward (factual) process that maps $\tilde{z}_0$ back to a reconstruction of the original $z_0$ under the original condition $c$.

The cycle-consistency loss encourages this round-trip process to recover the original latent:
\begin{equation}
   \ell_{\text{cycle}}(z_0, t, c, c') = \left\| z_0 - \hat{z}_0\big(\hat{z}_0(z_t, t, c'), t, c \big) \right\|.
\label{eq:cycle_loss}
\end{equation}

The full training objective combines three terms: (1) LDM loss in the counterfactual direction, (2) LDM loss in the factual direction (from the counterfactual back to the original condition), and (3) the cycle-consistency loss:
\begin{equation}
\mathcal{L}_\text{CDM} = \mathbb{E}_{z, \epsilon, t} \left[
\ell_{\text{LDM}}(\epsilon, z_t, t, c') + 
\ell_{\text{LDM}}(\epsilon, \tilde{z}_t, t, c) + 
\lambda \, \ell_\text{cycle}(z, t, c, c')
\right],
\label{eq:composite_loss}
\end{equation}
where $\tilde{z}_t$ is the noisy version of $\tilde{z}_0$ and $\lambda$ is a hyperparameter that controls the strength of the cycle-consistency regularization.

This composite loss encourages counterfactual samples to reflect the target condition while preserving anatomical structure from the source image.

\subsection{Training details}
% Note that directly minimizing Eq.~\eqref{eq:composite_loss} is computationally expensive, because it requires two sequential denoising steps to be done for every forward pass. We use checkpointing for saving memory to fit two forward passes in. Also, optimizing on a composite loss requires loss scheduling.
% In our experiments, we finetune the cycle-consistency loss in Eq.~\eqref{eq:cycle_loss} on a LDM pretrain with generation reconstruction loss, which we found works well in practice.
To ensure stable and efficient training, we adopt a two-phase strategy. We first pretrain our model using the standard LDM loss conditioned on demographic attributes (age, sex):
\[
\ell_{\text{LDM}}(\epsilon, z_t, t, c),
\]
which teaches the model to synthesize brain images under given conditions from noisy latents. We then finetune using our proposed composite loss in Eq.~\eqref{eq:composite_loss}. This phase introduces both counterfactual supervision and cycle-consistency, encouraging the model to make minimal and targeted changes when shifting between demographic conditions.

% \subsection{Results on Synthetic Experiments}

% \emph{Depending on the claim you make in the paper, this section may
%   not be relevant.}

% Especially if you are developing a new method, you will want to
% demonstrate its properties on synthetic or semi-synthetic experiments.
% Include experiments that will help us understand the contribution of
% the work to machine learning and healthcare. 

% \section{Results on Real Data} 

% \subsection{Results on Application A} 
\section{Experiments}
\subsection{Datasets and Experiment Setup}
%\subsubsection{Datasets}
We evaluate our cycle diffusion framework using a comprehensive dataset of 3D brain MRIs of control subjects, combined from four different studies. The total number of samples used is 27,066 and it includes:
\begin{itemize}
\item
Adolescent Brain Cognitive Development (ABCD) Study~\cite{Karcher2021}. 
    \item 
Alzheimer's Disease Neuroimaging Initiative (ADNI)~\cite{petersen2010alzheimer},
\item 
Human Connectome Project (HCP)~\cite{VANESSEN201362}, and 

\item
Parkinson's Progression Markers Initiative (PPMI)~\cite{Pulliam2011}. 
\end{itemize}
The dataset includes T1-weighted brain images with a resolution of $160 \times 192 \times 176$, which are skull-stripped and registered to MNI space.
All images have age and sex demographic metadata. 
To ensure robust training and evaluation, we partition the dataset into 21,051 volumes for training and 6,015 for validation. Given the age distribution skew between older and younger populations, we sample mini-batches such that each age group (binned by decade) is uniformly represented during training.

%\subsection{Experimental Setup.}
For direct image generation, we generate synthetic samples from a fixed set of conditions - age and sex - by linearly interpolating across ages from 5 to 100 years and ensuring a balanced sex distribution. 
For counterfactual generation, we selected 50 images from the validation dataset that vary linearly between ages 5 to 100 years. We generate counterfactual images that have converted ages with [$\pm$10, $\pm$30, $\pm$60], excluding ages that fall outside the range [0,100]. 
% \subsubsection{Experiment Setup}
\subsection{Baselines}
We compare against four generative models: two GAN-based models, (1) VAE-GAN and (2) $\alpha$-GAN~\cite{kwon2019braingan}, (3) an autoregressive model (AR) trained on VQVAE-based tokens~\cite{tudosiu2022autoregressive}, and (4) LDM model~\cite{pinaya2022brainimaginggenerationlatent}.
\begin{enumerate}
    \item 
VAE-GAN combines variational autoencoders with generative adversarial networks, leveraging the structured latent space of VAEs and the adversarial training of GANs to enhance output realism.
\item
$\alpha$-GAN~\cite{kwon2019braingan} is a hybrid of adversarial and variational objectives to improve mode coverage and uses the Wasserstein GAN with Gradient Penalty (WGAN-GP) loss \cite{gulrajani2017improvedtrainingwassersteingans} to lower training instability. 
%We also use $\alpha$-GAN to generated counterfactual images. Particularly, 
\item 
For autoregressive modeling, we evaluate a VQVAE-based tokenized autoregressive (AR) model~\cite{tudosiu2022autoregressive}. This approach compresses brain images into discrete latent codes using vector quantization, then trains a transformer network to autoregressively predict sequences of these tokens.
\item 
Finally, we include a LDM~\cite{pinaya2022brainimaginggenerationlatent} with details provided in Section \ref{sec:background}.
\end{enumerate}

All baseline and proposed models are conditioned on age and sex during training and sampling.
For GAN-based models, we condition on age and sex by appending them to the latent vector that is passed as input to the respective generators.
Note that GAN-based models support counterfactual generation by passing the same latent $z \sim p(z)$ with different appended conditions. 
% In particular, similar to the training process, the input image is encoded into the latent space by an encoder model. 
% In particular, instead of appending the true conditions of age and sex, we append the input counterfactual condition to this latent.
% Finally, the output counterfactual MRI is generated by the generator model using the latent and the counterfactual condition.

% Note that $\alpha$-GAN was originally proposed for slice-wise generation, but we reimplement it for 3D generation for better comparison.

\subsection{Training Setup and Evaluation Metrics}
Our diffusion model is based on a latent diffusion framework (LDM)~\cite{rombach2022highresolution}, with an encoder-decoder architecture and a diffusion U-Net backbone. The encoder consists of convolutional blocks with 3 levels of downsampling, and the number of latent channels is set to 8. For the diffusion model, the decoder reconstructs a single-channel 3D MRI volume. The UNet in our framework utilizes intermediate channel sizes of [384, 512, 512], with cross-attention applied to the latter two levels to incorporate conditioning data.

To train the LDM model, we use an L1 reconstruction loss and a patch-based adversarial loss to encourage realistic image generation. Additionally, a KL divergence penalty is employed to regularize the latent space. We set the weight for the adversarial loss to 0.005 and the KL penalty coefficient to 1e-7. For the cycle consistency loss, we use an L1 loss to minimize the difference between the real latent and the cycle-reconstructed latent. 
For LDM, we use a batch size of 1 and train the model for 400k iterations.
For CDM, we have a composite training loss (Equation \eqref{eq:composite_loss}), which requires loss scheduling. In practice, we find that pre-training on diffusion generation loss and then finetuning on cycle loss performs the best across all scheduling methods tested. We pre-trained the diffusion model on generation reconstruction loss for 400k iterations and then trained on cycle loss in Equation \ref{eq:composite_loss} for 2k iterations. All models are trained on an NVIDIA H100.

%\subsection{Metrics}
We evaluate our models using several metrics to assess image quality, conditioning adherence, and diversity. We use Frechet Inception Distance (FID)~\cite{heusel2018ganstrainedtimescaleupdate} to measure the quality of the generated images relative to real samples. The FID is computed using a 3D ResNet pre-trained on 23 medical imaging datasets~\cite{chen2019med3d}. %Lower FID scores indicate higher quality.

To measure diversity, we report the Multi-Scale Structural Similarity (MS-SSIM)~\cite{wang2003msssim} index. We calculated ms-ssim for all pairs of generated images, and average the value. Lower averaged ms-ssim means lower degree of image similarity and thus higher diversity. 

To evaluate the model's ability to preserve age and sex information, we use pretrained CNN-based models for age regression and sex classification. The error in these predictions quantifies how well the synthetic images retain the conditioning variables. 
The architectures for the age and sex predictors are CNNs with 4 downsampling levels and 2 convolutional blocks (conv, norm, and ReLU) per level.
The final layer is a channel-wise average pooling followed by a fully-connected layer down to a scalar output.
The age regressor is trained to minimize mean squared error (MSE) loss, while the sex classifier is trained to minimize binary cross-entropy loss.

\begin{table*}[t]
\centering
% \small
% \normalsize
\Huge
\resizebox{\textwidth}{!}{
\begin{tabular}{l|cccc|cccc}
\toprule%\hline
\textbf{ } & \multicolumn{4}{c|}{\textbf{Direct Generation}} & \multicolumn{4}{c}{\textbf{Counterfactual Age Generation}} \\ \cline{2-9} 
 & \textit{Age} & \textit{Sex} & \textit{FID} & \textit{MS-SSIM} & \textit{Age} & \textit{Sex} & \textit{FID} & \textit{MS-SSIM} \\ 
  & \textit{MAE} ($\downarrow$)~ & \textit{Acc} ($\uparrow$)~ & {$(\downarrow$)} & {$(\downarrow$)} & \textit{MAE} ($\downarrow$)~ & \textit{Acc} ($\uparrow$)~ & ($\downarrow$) & ($\downarrow$) \\ 
  \midrule %\hline
\textbf{Real}  & 2.23 & 88\% & - & 0.75 & 2.23 & 88\% & - & 0.75 \\
% \textbf{Random}  & 30.45 & 50\% & - & - & 30.45 & 50\% & - & - \\ 
\hline
\textbf{VQVAE-AR}  & 40.37 & 46\% & 55.30 & 0.81 & - & - & - & - \\
\textbf{VAE-GAN}  & 28.57 & 52\% & 93.29 & 0.91 & - & - & - & - \\
\textbf{$\alpha$-GAN}  & 33.30 & 50\% & 60.19 & 0.97 & 36.86 & 48\% & 191.55 & 0.93 \\
\textbf{LDM}  & 18.80 & 84\% & 35.50 & 0.80 & 9.71 & 82\% & \textbf{90.12} & 0.822 \\
\textbf{CDM}  & \textbf{15.39} & \textbf{88\%} & \textbf{35.46} & \textbf{0.79} & \textbf{7.87} & \textbf{86\%} & 91.86 & \textbf{0.8047} \\
\bottomrule %\hline
\end{tabular}
}
\caption{Comparison of models for direct generation and counterfactual age generation. Lower values in Age MAE, FID, and MS-SSIM, and higher values in Sex Accuracy indicate better model performance.The CDM model achieves the best results in  both generation and counterfactual tasks, outperforming others in condition adherence(age MAE, sex accuracy), image quality(FID) and image diversity(MS-SSIM).}
\label{tab:comparison}
\end{table*}

% \subsection{Results}
% \begin{figure}
% \includegraphics[width=\textwidth]{images/add 3 baseline.png}
% \caption{Mean absolute error of age prediction for direct synthetic generation for VQVAE with auto-regressive training, W-Gan, VAE-GAN, LDM and CDM. MAE of CDM out perform other baseline in most age ranges. } \label{age direct}
% \end{figure}

\begin{figure}[t]
\includegraphics[width=\textwidth]{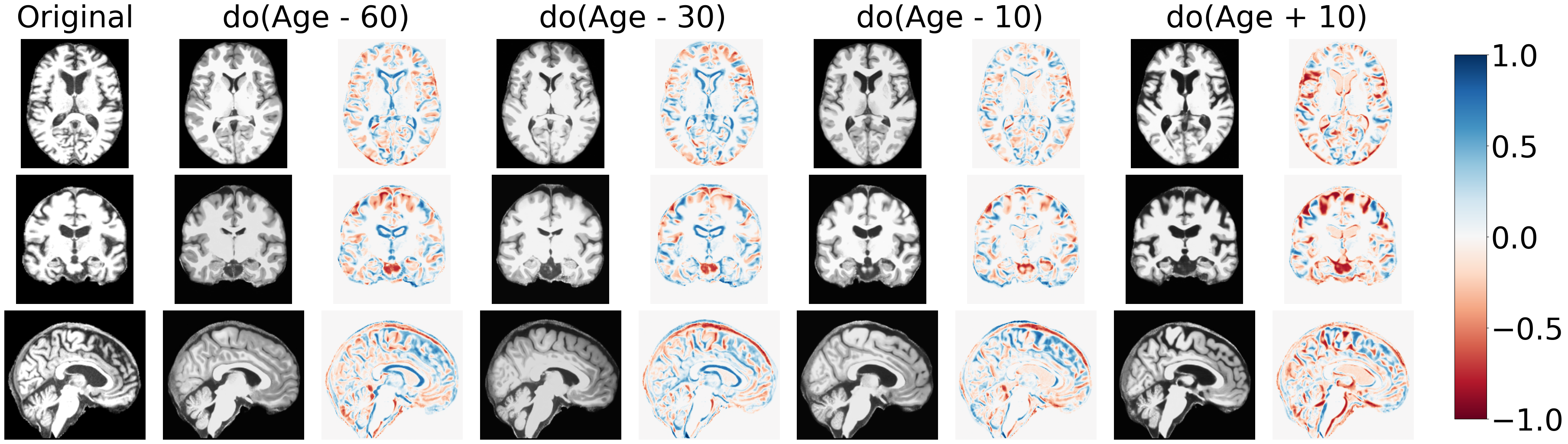}
\caption{Representative sample along three views of a 79-year-old female brain (left column) along with four counterfactuals and their difference maps with respect to the original sample: ($-60$, $-30$, $-10$, $+10$). Sex is unchanged. As age decreases, ventricle size shrinks (indicated by blue in difference maps) and the cortical surface appears thicker, reflecting reduced brain atrophy. In contrast, the older counterfactual shows enlarged ventricles (indicated by red in difference maps) and increased cortical thinning, consistent with age-related brain atrophy and neuro-degeneration \cite{gur1995sex,Xu112,fjell2009minute}. 
These gradual structural changes demonstrate the model's ability to realistically capture age-related brain morphology across different age intervals.
}
\label{cf_diff}
\end{figure}

\subsection{Direct Generation}
Figure \ref{direct visualization} depicts results for direct generation, where each image is synthesized directly using a specified age and sex condition 
(e.g., age=13, sex="male"). 
% The generated images show realistic brain MRI with well-preserved anatomical structures. % and clear age-related differences between the generated samples. 
In comparison to the baseline methods, CDM synthesizes images with finer structural details and sharper boundaries, particularly noticeable in the cortical folds and sulcal regions.
In contrast, the GAN-based samples are blurry and lack anatomical detail. 
VQVAE-AR produces sharper images, but cortical folds are unresolved.
LDM produces the sharpest boundaries out of all baselines, but lacks definition in the ventricles and folds.
%We attribute the visual realism in CDM samples to the more restrictive generations enforced by the cycle-consistency constraint. 
% boundaries but it struggles at anatomical plausibility. 
% These differences are critical in clinical settings where precise anatomical accuracy is essential for applications such as brain disease progression or surgical planning. 

% Our improvement can also be shown through quantitative result in Table \ref{tab:comparison}. 
The left side of Table \ref{tab:comparison} quantifies the metrics on direct generation. 
We observe that CDM has better performance across all metrics. 
Notably, samples generated by CDM have the lowest age MAE and highest sex accuracy; we hypothesize that the cycle-consistency constraint imposes additional regularization that improves the adherence of the generator to the input conditions. 
% image quality (measured by FID) and image diversity(measured by MS-SSIM). 
% Being able to perform better than other baselines in both tasks shows the generalizability of our method. 

% In Table \ref{tab:ldm_comparison}, we show that the age mae for counterfactual generation is also the best for across most counterfactual changes which shows that our method generalizes across both direct and counterfactual generations.

\subsection{Counterfactual Generation}
In Figure \ref{cf_diff}, we depict a representative generated sample of a 79-year-old female along with several counterfactual samples of the original sample at ages 19, 49, 69, and 89.
% where the model transforms a brain MRI from one age condition to another age condition ($\pm$ 10, $\pm$ 30) while preserving the original sex. 
For age-reduced counterfactuals (age-10, age-30, age-60), the size of the ventricle decreases, represented by a blue color in the ventricle region, which corresponds to a smaller ventricle in younger individuals. 
Conversely, for the age-increased counterfactual (age+10), the model simulates an enlargement of the ventricles. 
These changes align with known age-related ventricular enlargement observed in real data and in prior work~\cite{gur1995sex,Xu112,fjell2009minute}. 
Additionally, cortical changes are observed: the younger brain exhibits a fuller, more robust cortex, while the older brain shows noticeable cortical atrophy, which is consistent with real-life cortical thinning due to aging. 
These results indicate that the model effectively captures key structural aging patterns in brain anatomy.

Tables~\ref{tab:comparison} and \ref{tab:counterfactual_age_mae} quantify the age MAE metric for all counterfactuals.
The right side of Table~\ref{tab:comparison} quantifies metrics aggregated across all counterfactuals, where we observe that counterfactuals generated by CDM outperform baseline counterfactuals across all metrics except for FID, where CDM is on par with LDM. 
Table~\ref{tab:counterfactual_age_mae} further breaks down counterfactual metrics across different counterfactual groups corresponding to the age deltas.
We observe that CDM has the lowest age MAE across nearly all counterfactual groups.
% this can be shown by Table \ref{tab:counterfactual_age_mae}, where we measure the age prediction mean absolute error of counterfactual generation for six age changes($\pm 10$, $\pm 30$ and $\pm 60$). 
% Across Our method consistently outperforms baseline methods for age conditioning adherence. Overall, our method generalizes for both types of generations and is more robust in almost all age ranges. 

% TODO : add ventrical, figure 4 counterfactual generation. blue and red. Hypotheses. We hypothesize. in figure 3 , our model exhibit better, we hypothesize this is because. At age is worse prevalence, because our model is better regularize, and the data efficient. Justify in the experiment. 

% \setlength{\tabcolsep}{12pt}  % Adjust the column spacing here

\begin{table}[ht]
\setlength{\tabcolsep}{1pt} 
\centering
\begin{tabular}{l|cccccc}
\toprule
 & do(age-60) & do(age-30) & do(age-10) & do(age+10)  & do(age+30)  & do(age+60)  \\
\midrule
$\alpha$-GAN\cite{kwon2019braingan}        & 7.38  & 18.96  &  28.18 & 42.78  & 52.97  & 67.86  \\
LDM\cite{pinaya2022brainimaginggenerationlatent}        & 3.75  & \textbf{8.70}  & 6.56  & 9.88  & 15.09  & 12.83  \\
CDM (ours) & \textbf{3.35}  & 10.03  & \textbf{6.14}  & \textbf{8.62}  & \textbf{9.97}  & \textbf{7.72}  \\
\bottomrule
\end{tabular}
\caption{Comparison of the Mean Absolute Error (MAE) for age-related counterfactual image generation using three models: $\alpha$-GAN, LDM, and CDM. The models are evaluated on their ability to generate images at different counterfactual ages ($\pm 10$, $\pm 30$, $\pm 60$ years) from the baseline age. "Do" means applying age change counterfactual generation on the image. Counterfactual ages that are not the in range [0,100] are discarded. Lower MAE values indicate better performance in generating realistic age-altered images, with CDM consistently outperforming both $\alpha$-GAN and LDM across most age offsets. It shows that cycle consistent training effectively capture semantic level feature like age. }
\label{tab:counterfactual_age_mae}
\end{table}
% \section{Downstream Application} 
% \subsection{Data augmentation}
% We test the model's ability to generate synthetic images to improve the performance of a downstream application-sex classification. We use 3D-Resnet50 for sex classification. The model is trained with learning rate 1e-5 for 20 epochs for all sex classification experiments. We compared the performance of 6 different training datasets:  1. 200 real image, 2.200 synthetic image from CDM direct generation, 3. 200 synthetic image from CDM counterfactual generation,  4. 200 real image with 200 synthetic image from CDM direct generation. 5. 200 real image with 200 synthetic image from CDM counterfactual generation. 6. 200 real image with 200 synthetic image from CDM direct generation and 200 synthetic image from CDM counterfactual generation. 
% \begin{table}[h]
%     \centering
%     \begin{tabular}{lcccccc}
%         \toprule
%         \textbf{Training Data} & \textbf{Acc.} & \textbf{AUC} & \textbf{F1} & \textbf{Sens.} & \textbf{Spec.} \\
%         \midrule
%         200 real images & & & & & \\
%         200 CDM direct & & & & & \\
%         200 CDM counterfactual  & & & & & \\
%         200 real+200 CDM direct & & & & & \\
%         200 real+200 CDM counterfactual & & & & & \\
%         200 real+200 CDM direct+200 CDM counterfactual & & & & & \\
%         \bottomruled
%     \end{tabular}
%     \caption{Performance metrics for different training augmentation. The metric for }
%     \label{tab:results}
% \end{table}
% \FloatBarrier

\section{Discussion and Conclusion} 

% Our proposed Cycle Diffusion Model demonstrates significant improvements in both direct and counterfactual brain MRI generation compared to baseline approaches. The cycle training framework not only enhances the visual quality of generated images but also significantly improves conditioning adherence, which is crucial for medical applications where precise control over demographic factors is essential.

\noindent\textbf{Technical Contributions and Implications: } The cycle-consistency constraint provides a novel approach to improving conditional image generation for medical imaging. Our framework explicitly enforces conditioning adherence through a bidirectional training process, allowing the model to learn minimal changes necessary while maintaining anatomical plausibility. Our results show that CDM achieves the best conditioning adherence across most generation tasks with \emph{improved} image quality. This improved performance can be attributed to cycle consistency, which forces the model to learn more precise, condition-specific transformations.

\noindent\textbf{Clinical and Healthcare Applications:}
CDM enables targeted augmentation of under\mbox{--}represented groups, likely enhancing fairness and robustness of downstream diagnostic models. It can also simulate disease trajectories: generating counterfactual brain images at different ages lets clinicians visualize how a disorder might evolve for a given patient and plan earlier interventions.

\noindent\textbf{Limitations:}
Although CDM demonstrates promising performance in medical image condition translation tasks, there are still several limitations to be addressed. First, due to the multi-step sampling process inherent to diffusion models, CDM incurs relatively high computational costs during inference, which limits its applicability in scenarios where high efficiency is required. Second, the model exhibits performance fluctuations when the training data are limited or when there is a significant distribution shift between the source and target domains, indicating its sensitivity to data distribution and the need for improved generalization. Finally, this study primarily focuses on the development and validation of the technical approach, without systematic evaluation in real-world clinical tasks or applications. Our promising results open up opportunities for further exploration of CDM’s clinical utility and real-world impact.

\section*{Acknowledgment}
This work was supported in part by National Institutes of Health grant AG089169 and the Stanford Institute for Human-Centered AI (HAI) Hoffman-Yee Award. 

\bibliographystyle{splncs04}
\bibliography{sample}
% \input{samplepaper.bbl}




% LNCS Homepage, \url{http://www.springer.com/lncs}. Last accessed 4
% \end{thebibliography}

\appendix
% \addcontentsline{toc}{section}{Appendix}
\section{Preliminary} \label{sec:background}
%\subsection{Latent diffusion model}
Our proposed Cycle Diffusion Model (CDM) extends the Latent Diffusion Model (LDM)~\cite{rombach2022highresolution}.
An LDM is trained in two stages. 
First, an autoencoder learns to map images \( x \) to latent embeddings \( z \) using an encoder \( \mathcal{E} \) and decoder \( \mathcal{D} \). 
Second, a conditional diffusion model $\epsilon_\theta$ is trained on the latent space. 
Let $t \in \{1, ..., T\} = [T]$ denote the timestep index.
During training, the model learns to denoise a diffusion process by predicting the noise \( \epsilon \) added to the latent at each timestep $t$. 
The diffusion process is a Markov process with Gaussian transitions parameterized by a decreasing sequence $\alpha_{[T]} \in (0,1]^T$, such that:
% In the forward process, time-dependent noise is added to $z_0$:
\begin{equation}
    z_t = \sqrt{\alpha_t} z_0 + \sqrt{1-\alpha_t} \epsilon, \ \text{where} \ \epsilon \sim \mathcal{N}(0, I).
\end{equation}
The denoising network \( \epsilon_{\theta} \), typically implemented as a time-conditional U-Net, learns to predict the noise conditioned on the timestep $t$ and the conditioning information $c$:
% The LDM training loss is:
\begin{equation}
\ell_{\text{LDM}} = \mathbb{E}_{z, \epsilon \sim \mathcal{N}(0,1), t} \left[\| \epsilon - \epsilon_{\theta}(z_t, t, c) \|_2^2\right] := \mathbb{E}_{z, \epsilon \sim \mathcal{N}(0,1), t} \left[\ell_{LDM}(\epsilon, z_t, t, c) \right].
\end{equation}

To generate a sample, the model starts with a noisy latent representation \( z_T \) sampled from \(\mathcal{N}(0, I) \) along with a given condition $c$ and performs iterative denoising using $\epsilon_\theta$ to recover a clean latent \( \hat{z}_0 \). 

\subsection{DDIM Sampling and Inversion}

Denoising diffusion implicit model (DDIM)-based inversion is a popular method for generating counterfactuals using diffusion models and leverages the deterministic nature of DDIM sampling~\cite{alaya2024mededit,jeanneret2022diffusionmodelscounterfactualexplanations,Song2020DenoisingDI}.
The deterministic DDIM sampling step is:
\begin{equation}
    z_{t-1} = \sqrt{\alpha_{t-1}} \underbrace{\left(\frac{z_t - \sqrt{1-\alpha_t} \epsilon_\theta(z_t, t, c)}{\sqrt{\alpha_t}} \right)}_{\hat{z}_0(z_t, t, c)}+ \sqrt{1-\alpha_{t-1}} \cdot \epsilon_\theta(z_t, t, c).
\end{equation}
Note that $\hat{z}_0(z_t, t, c)$ is the estimate for the original sample $z_0$ at the current timestep $t$.

One can invert this sampling procedure to obtain $z_t$ as a function of $z_{t-1}$:

\begin{equation}
z_t = \frac{\sqrt{\alpha_t}}{\sqrt{\alpha_{t-1}}} z_{t-1} + \left(\sqrt{1 - \alpha_t} - \frac{\sqrt{\alpha_t} \sqrt{1 - \alpha_{t-1}}}{\sqrt{\alpha_{t-1}}}\right) \epsilon_\theta(z_t, t, c),
\label{eq:zt}
\end{equation}
By inverting for all $t \in [T]$, we can obtain the original noisy latent $z_T$,
% where \( \epsilon_\theta(z_t, t, c) \) represents the noise predicted by the model at each timestep \( t \) for the latent variable \( z_t \), conditioned on the metadata \( c \).
and perform forward sampling conditioned on new metadata \( c' \) to regenerate the image with the desired counterfactual condition.

\end{document}